\documentclass[conference]{IEEEtran}
\IEEEoverridecommandlockouts
\usepackage{cite}
\usepackage{amsmath,amssymb,amsfonts}
\usepackage{algorithmic}
\usepackage{graphicx}
\usepackage{textcomp}
\usepackage{xcolor}
\usepackage{csquotes}
\usepackage{siunitx}
\usepackage{algorithm2e}
\usepackage{tikz}
\usepackage{pgfplots}
\usepackage{url}
\pgfplotsset{compat=1.13}

\usepackage[textsize=tiny]{todonotes}

\def\BibTeX{{\rm B\kern-.05em{\sc i\kern-.025em b}\kern-.08em
    T\kern-.1667em\lower.7ex\hbox{E}\kern-.125emX}}
    
\usepackage[firstpage=true,color=black,placement=top,scale=1,opacity=1,vshift=-0.3cm]{background}
\SetBgContents{
 \begin{minipage}{2.3\linewidth}
 \small
\textcopyright 2022 IEEE.  Personal use of this material is permitted.  Permission from IEEE must be obtained for all other uses, in any current or future media, including reprinting/republishing this material for advertising or promotional purposes, creating new collective works, for resale or redistribution to servers or lists, or reuse of any copyrighted component of this work in other works. A definitive version will be published in \textbf{5th IEEE International Conference on Image Processing Applications and Systems 2022 (IPAS).}
 \end{minipage}
 }
\begin{document}

\title{Cell tracking for live-cell microscopy using an activity-prioritized assignment strategy\\
\thanks{This work was performed as part of the Helmholtz School
for Data Science in Life, Earth and Energy (HDS-LEE) and received funding from the Helmholtz Association and Deutsche Forschungsgemeinschaft (DFG, German Research Foundation) $333849990/\text{GRK}2379$ (IRTG Modern Inverse Problems)
.
}
}
\makeatletter
\newcommand{\linebreakand}{%
  \end{@IEEEauthorhalign}
  \hfill\mbox{}\par
  \mbox{}\hfill\begin{@IEEEauthorhalign}
}
\makeatother
\author{\IEEEauthorblockN{Karina Ruzaeva}
\IEEEauthorblockA{\textit{AICES Graduate School} \\
\textit{RWTH Aachen}\\
Aachen, Germany \\
\textit{IBG-1: Biotechnology} \\
\textit{Forschungszentrum Jülich GmbH}\\
Jülich, Germany \\
ruzaeva@aices.rwth-aachen.de}
\and
\IEEEauthorblockN{Jan-Christopher Cohrs}
\IEEEauthorblockA{\textit{AICES Graduate School} \\
\textit{RWTH Aachen}\\
Aachen, Germany \\
cohrs@aices.rwth-aachen.de}
\and
\IEEEauthorblockN{Keitaro Kasahara}
\IEEEauthorblockA{\textit{IBG-1: Biotechnology} \\
\textit{Forschungszentrum Jülich GmbH}\\
Jülich, Germany \\
kasahara@fz-juelich.de}
\linebreakand
\IEEEauthorblockN{Dietrich Kohlheyer}
\IEEEauthorblockA{\textit{IBG-1: Biotechnology} \\
\textit{Forschungszentrum Jülich GmbH}\\
Jülich, Germany \\
d.kohlheyer@fz-juelich.de}
\and
\IEEEauthorblockN{Katharina Nöh}
\IEEEauthorblockA{\textit{IBG-1: Biotechnology} \\
\textit{Forschungszentrum Jülich GmbH}\\
Jülich, Germany \\
k.noeh@fz-juelich.de}
\and
\IEEEauthorblockN{Benjamin Berkels}
\IEEEauthorblockA{\textit{AICES Graduate School} \\
\textit{RWTH Aachen}\\
Aachen, Germany \\
berkels@aices.rwth-aachen.de}
}

\maketitle

\begin{abstract}
Cell tracking is an essential tool in live-cell imaging to determine single-cell features, such as division patterns or elongation rates.
Unlike in common multiple object tracking, in microbial live-cell experiments cells are growing, moving, and dividing over time, to form cell colonies that are densely packed in mono-layer structures. With increasing cell numbers, following the precise cell-cell associations correctly over many generations becomes more and more challenging, due to the massively increasing number of possible associations.

To tackle this challenge, we propose a fast parameter-free cell tracking approach, which consists of activity-prioritized nearest neighbor assignment of growing cells and a combinatorial solver that assigns splitting mother cells to their daughters. As input for the tracking, Omnipose is utilized for instance segmentation. Unlike conventional nearest-neighbor-based tracking approaches, the assignment steps of our proposed method are based on a Gaussian activity-based metric, predicting the cell-specific migration probability, thereby limiting the number of erroneous assignments.
In addition to being a building block for cell tracking, the proposed activity map is a standalone tracking-free metric for indicating cell activity.
Finally, we perform a quantitative analysis of the tracking accuracy for different frame rates, to inform life scientists about a suitable (in terms of tracking performance) choice of the frame rate for their cultivation experiments, when cell tracks are the desired key outcome.
\end{abstract}


\section{Introduction}
Live-cell (time-lapse) microscopy combined with microfluidic lab-on-chip technology enables observing single-cell features, such as cell length distributions and elongation rates, with spatio-temporal resolution in 2D~\cite{Grnberger:540646}. 
Unlike standard tracking tasks (i.e., high-frame rate people or car tracking),
tracking living microorganisms has specific, biology-related challenges. However, considering the height of the cultivation chamber (about the size of the cell's width), cell occlusions are prevented, while in standard tracking tasks, e.g., tracking of people, objects occlusions are almost inevitable.

Generally, at the beginning of a live-cell image sequence, only few cells are present and those are sparsely distributed. In this situation, tracking, i.e., linking cells over frames including the detection of division events, is relatively straightforward. As the experiment progresses, it becomes more and more challenging to detect the correct associations, in particular in exponential growth regimes. In crowded cell colonies, with large numbers of divisions, which often cause immense and random cell displacements between consecutive frames, identifying cell associations correctly, becomes hard even for experts~\cite{Theorell_2018}.

Compared to standard object tracking tasks with a high frame rate, where the object positions can be predicted quite accurately because of relatively small frame-to-frame object displacements \cite{Cazzolato2018}, in microbial live-cell imaging, researchers have to deal with notoriously low frame-rates, relative to the cell doubling times. 
The low frame rate in such time series generally cannot be increased since illumination from live-cell imaging, when coupled with fluorescence, leads to phototoxic effects that may change the cells' growth behavior, thereby entailing the risk to distort the interpretation of the results. Consequently, without prior information on the expected division timing and splitting behavior of the cells, the inevitably low frame rates together with chaotic cell movements %
are leaving us dealing with many daughter cells that need to be linked to their mothers.

To overcome these challenges, we propose a fast parameter-free tracking approach that uses ML-based segmentation as input. The tracking is based on activity-prioritized adaptive nearest neighbor linking and a combinatorial mother-daughters assignment solver, with only one term in the linking loss function. This combinatorial assignment solver uses the biologically motivated constraint that one mother is assigned to (at most) two daughters, provided that the frame rate is higher than the division rate. 

\subsection{Related work} 
In the literature, cell tracking methods are split into unsupervised methods, such as tracking by model evolution and tracking by detection; and supervised, usually ML-based, approaches. 

In tracking by model evolution approaches, an initial segmentation is propagated over time, meaning that the result from the previous frame becomes an initialization for the next one, thereby performing simultaneous segmentation and tracking. These methods require a considerable overlap between the initialized contour and an object to be segmented, which may not be the case in the low-frame rate datasets. Besides, taking into account the variational nature of these methods, they may be computationally expensive for large cell colonies \cite{Maska2013, Dzyubachyk_2010, Arbelle2018}, making the algorithms slow compared to tracking by detection algorithms.

Tracking by detection methods split the task into detection, often in form of instance segmentation, and tracking. The segmentation is performed with a variety of available cell segmentation methods \cite{Cutler2021, Stringer2020, Ruzaeva2022}.
These tracking methods link the segmented cells between consecutive frames based on their similarities, i.e., by finding correspondences between cell features in successive frames. Here, cell association becomes complicated when the feature similarity of a cell to its within-frame neighbors is comparable to the similarity of the same cell in consecutive frames.
Most traditional cell tracking association strategies are performed by comparing the feature vectors, where the dominant feature typically is the cell location, i.e., so-called nearest neighbor tracking \cite{Ulman_2017}. In addition to the cell location, other features are utilized, e.g., spectral features of single-cell image crops \cite{Cuny2021}.

Tracking by detection is performed by linking segmentation masks over time based on a loss, i.e., the \enquote{linking} measure. A simple loss is the Euclidean distance between the positions of the cell centroids. To prevent non-physiological long-distance associations, a maximal distance limit is often applied to the distances \cite{Tinevez_2017}.
Other linking measures are based on handcrafted features, such as morphology and features of the cell's neighborhood \cite{Theriault_2011,Balomenos_2015,Sarmadi_2022}. The loss, which consists of several terms, requires the proper, manually tuned weighting of the loss components. That makes the algorithm hard to generalize across microorganisms featuring different growth behaviors. Besides, the cell daughters may lose the mother's features, such as the orientation angle, after the division, making the assignment problems less trivial.

As an alternative to the feature-vector comparison, some methods utilize cross-correlation \cite{L_ffler_2021}. The cross-correlation-based linking shows promising results for cells with more complex morphology, but the approach is not optimized for cases, where cells look very much alike, e.g. coccoid or rod-shaped microorganisms.

To account for possible erroneous associations and segmentation imperfections, to improve the obtained cell lineage, some works propose to combine the global linking with segmentation refinement \cite{Magnusson2015, L_ffler_2021}. But indeed, and with current improvements in the segmentation tools, the refinement may be counter-productive (shown in the Results) and even become a source of errors.
Rather than aiming at deriving a single lineage, Uncertainty-Aware Tracking  relies on a Bayesian approach and keeps track of all possible lineages, ranking them by their probability, shows promising results for low-frame rate datasets \cite{Theorell_2018}. However, the approach is computationally expensive, extremely space demanding, and, thus, unsuitable for the analysis of big cultivation datasets, like \cite{Schito2022}.

Unsupervised cell tracking approaches are still dominant in the cell tracking field. Although, supervised ML-based cell tracking approaches exist, e.g., \cite{Loeffler2022, Lugagne2020}. These methods require training data, and considering every microfluidic experiment's uniqueness (different microorganisms, microscopy setup, etc.), training and benchmark datasets are rare, and manual tracking annotations are extraordinarily laborious to produce.

\section{Target microorganisms and cultivation details}

To assess the performance of the method, the tool is tested with in-house generated test dataset of two widely used microbacteria \textit{Corynebacterium glutamicum} (\textit{C.~glutamicum}) and \textit{Escherichia coli} (\textit{E.~coli}). 
\paragraph{E.~coli}
In biotechnology, the gram-negative bacterium \textit{E.~coli} is a well-studied model organism and an expression host for large-scale production of proteins \cite{Lee2009}.
The bacterium is rod-shaped, and about \SIrange{1}{2}{\micro\meter} long and \SI{0.5}{\micro\meter} wide. \textit{E.~coli} divides uniformly, "diffusively" elongating along the cell main axis and with a cell division occurring in the middle \cite{Grnberger:540646}.
To test our approach, the wild type strain \textit{E.~coli} MG1655 was used.
The strain was cultured under anaerobic conditions in a microfluidic device for microbial single-cell analysis, coupled with time-lapse microscopy.
Experiments were carried out using an inverted time-lapse live cell microscope equipped with a 100×oil immersion objective and a temperature incubator. Phase contrast images of the growing microcolonies were captured every five minutes. The observed average single-cell division time is $78\pm24$ min (average time between a cell's birth and its division, calculated from the ground truth lineages). For further cultivation details, we refer to \cite{Kaganovitch_2018}.

\paragraph{C.~glutamicum}
\textit{C.~glutamicum} is a gram-positive microorganism, used in industrial biotechnology  for the production of amino acids, especially L-glutamate and L-lysine~\cite{Eggeling2005}.
As \textit{E.~coli}, \textit{C.~glutamicum} has a rod-shaped cell geometry, and typical strains range from cell lengths of \SI{2}{\micro\meter} to \SI{5}{\micro\meter}, while the width is about \SI{1}{\micro\meter}~\cite{Grnberger:540646}.
In contrast to the symmetric cell division behavior of \textit{E.~coli}, \textit{C.~glutamicum} shows apical growth (i.e., the cell wall is expanding at the cell poles), and tends to divide asymmetrically into two unequal daughters. In addition, the peculiar dynamic \enquote{V-snapping} of \textit{C.~glutamicum} cells after division (Fig.~\ref{fig:comparison}) \cite{Zhou2019}, complicates the tracking task, especially when it comes to finding correspondences between consecutive frames, when the majority of cells in a population are dividing at the same time, pushing the cells in their neighborhood.
In our numerical experiments, the wild type strain \textit{C.~glutamicum} ATCC 13032 was used. The imaging parameters of the experiment are similar to the \textit{E.~coli} cultivation experiment. However, unlike in the \textit{E.~coli} experiment, phase contrast images here were captured every two minutes. The observed average single-cell division time is $79\pm13$ min. For further cultivation details, we refer to \cite{Mosheiff}. 

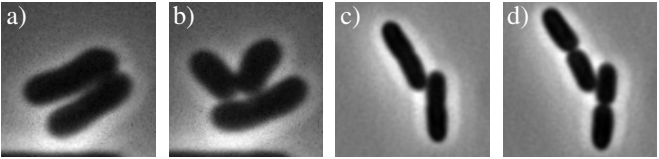
\begin{figure}[!htb]
\centering
\begin{tikzpicture}
  \node[anchor=south west,inner sep=0] (Bild) at (0,0)
    {\includegraphics[width=.23\linewidth]{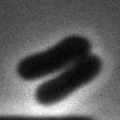}};
  \begin{scope}[x=(Bild.south east),y=(Bild.north west)]
    \node[color=white] at (0.1,0.9) {a)};
  \end{scope}
\end{tikzpicture}
\begin{tikzpicture}
  \node[anchor=south west,inner sep=0] (Bild) at (0,0)
    {\includegraphics[width=.23\linewidth]{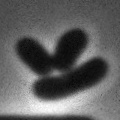}};
  \begin{scope}[x=(Bild.south east),y=(Bild.north west)]
    \node[color=white] at (0.1,0.9) {b)};
  \end{scope}
\end{tikzpicture}
\begin{tikzpicture}
  \node[anchor=south west,inner sep=0] (Bild) at (0,0)
    {\includegraphics[width=.23\linewidth]{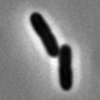}};
  \begin{scope}[x=(Bild.south east),y=(Bild.north west)]
    \node[color=white] at (0.1,0.9) {c)};
  \end{scope}
\end{tikzpicture}
\begin{tikzpicture}
  \node[anchor=south west,inner sep=0] (Bild) at (0,0)
    {\includegraphics[width=.23\linewidth]{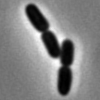}};
  \begin{scope}[x=(Bild.south east),y=(Bild.north west)]
    \node[color=white] at (0.1,0.9) {d)};
  \end{scope}
\end{tikzpicture}
\caption[]{Illustration of \textit{C.~glutamicum} (a-b) and \textit{E.~coli} (c-d) division events.}
\label{fig:comparison}
\end{figure}

\section{Methods}
We propose a new nearest neighbor tracking approach, which includes an \enquote{activity}-based prioritization and loss function measure, and a combination of a minimal loss single-assignment problem with Jonker-Volgenant \cite{Crouse2016} linear assignment strategy. The proposed method is efficient and robust, does not require any parameter tuning, and generalizes to many microorganisms, especially to those with low to none motility.

\subsection{Segmentation and preprocessing}
\label{seg}
As segmentation framework, we use the recent U-net-based tool Omnipose \cite{Cutler2021}. The Omnipose network was trained on similarly looking phase contrast images of bacteria and addresses our segmentation problem well. We use the default weights and the \texttt{bact\_omni} model type. 
We did not observe any false negative segmentation instances. The few false positives segmentation instances were filtered out by setting a known minimal cell area limit, specific to the microorganism.

\subsection{Activity map}
We propose a novel measure for cell activity that we call \enquote{activity map} (AM), highlighting the objects/cells that are likely to experience translation, division, or tilting.

The activity of cells is typically derived from tracking results. However, the approximate estimation of the cell activity, as will be shown, is possible without tracking information and may even enhance the tracking. The proposed metric is based on the intensity differences of consecutive frames combined with the cells' segmentation masks, obtained in the segmentation step. The calculation of intensity difference of two consecutive frames was utilized by \cite{Sarmadi_2022} as a preliminary step for the cell tracking to estimate the cells' motion vectors, but limited to high frame rate (less than five minutes) datasets.

To compute the proposed AM, we first calculate the moving pixel-wise standard deviation ($S$) over a stack of frames, i.e.,
\begin{equation}
S_{t}=\sqrt{\frac{1}{n_{-} + n_{+}+1}\sum_{s=t-n_-}^{t+n^+}\left(I_{s}-\overline{I_{t-n_{-}},...,I_{t+n_{+}}}\right)^2},
\end{equation}
where $n_{-},n_{+}\in\mathbb{N}$, $I_s$ for $s\in\{1,\ldots,N\}$ is the $s$-th image (frame) in the image stack, $N$ is the total number of frames, and $\overline{I_{t-n_{-}},...,I_{t+n_{+}}}$ denotes the average of the image stack over frames $I_{t-n_{-}}$ to $I_{t+n_{+}}$. For the border regions, the interval is taken from $\max(t-n_{-},1)$ to $\min(t+n_{+},N)$.
The window for the calculation is specified by the user and can be tailored for the application. For general purpose visualization, we suggest taking the symmetric window ($n_{-}=n_{+}$) from $t_{t-n_{-}}$  to $t_{i+n_{+}}$ for $n_{-}<t<N-n_{+}$.

To calculate $a_{t,i}$, the activity of $i$-th cell at time $t$, we integrate the obtained map $S_t$ over the corresponding segmentation mask. To reduce the influence of the cell area, i.e., bigger cells would have more activity, the value was normalized by dividing by the area, i.e.,
\begin{equation}
a_{t,i}=\frac{\sum_{x=0,y=0}^{X,Y} M_{t,i}(x,y) S_{t}(x,y)}{\sum_{x=0,y=0}^{X,Y} M_{t,i}(x,y)}
\end{equation}
where $M_{t,i}$ is the binary segmentation mask for the $i$-th segmentation instance of the $t$-th frame, cf. \ref{seg}.
As a prioritization and activity metric for cell tracking, we consider only two ($n_{-}=0, n_{+}=1$) consecutive frames.
The proposed activity map is illustrated in Fig.~\ref{fig:actmap}, where each cell's segmentation mask was \enquote{colored} with the respective activity value.
\begin{figure}
\centering
\begin{tikzpicture}
  \node[anchor=south west,inner sep=0] (Bild) at (0,0)
    {\includegraphics[width=.46\linewidth]{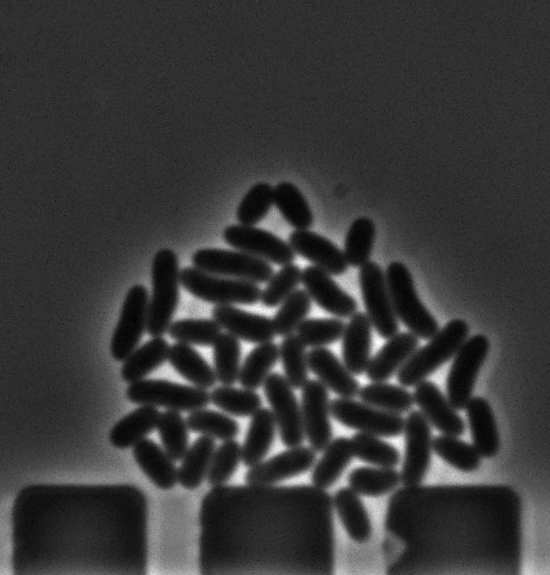}};
  \begin{scope}[x=(Bild.south east),y=(Bild.north west)]
    \node[color=white] at (0.075,0.925) {a)};
  \end{scope}
\end{tikzpicture}
\begin{tikzpicture}
  \node[anchor=south west,inner sep=0] (Bild) at (0,0)
    {\includegraphics[width=.46\linewidth]{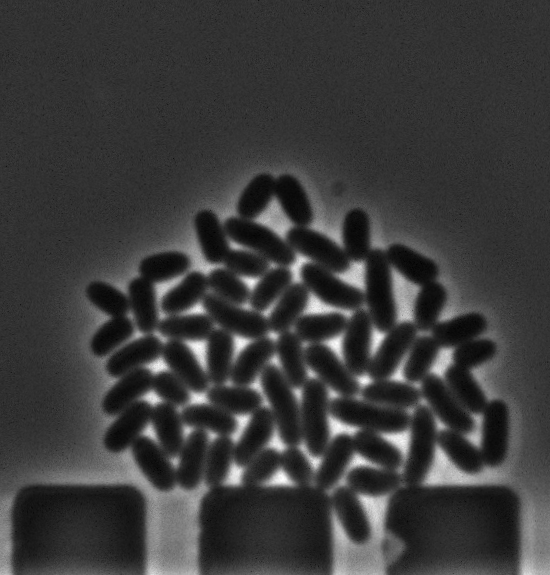}};
  \begin{scope}[x=(Bild.south east),y=(Bild.north west)]
    \node[color=white] at (0.075,0.925) {b)};
  \end{scope}
\end{tikzpicture}
\\[0.75ex]
\begin{tikzpicture}
  \node[anchor=south west,inner sep=0] (Bild) at (0,0)
    {\includegraphics[width=.46\linewidth]{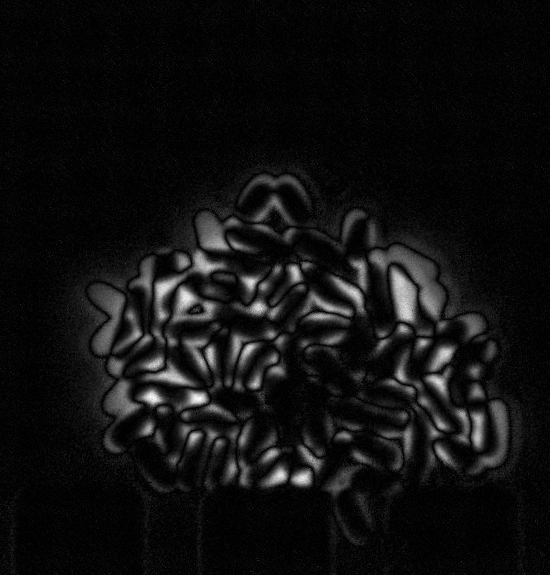}};
  \begin{scope}[x=(Bild.south east),y=(Bild.north west)]
    \node[color=white] at (0.075,0.925) {c)};
  \end{scope}
\end{tikzpicture}
\begin{tikzpicture}
  \node[anchor=south west,inner sep=0] (Bild) at (0,0)
    {\includegraphics[width=.46\linewidth]{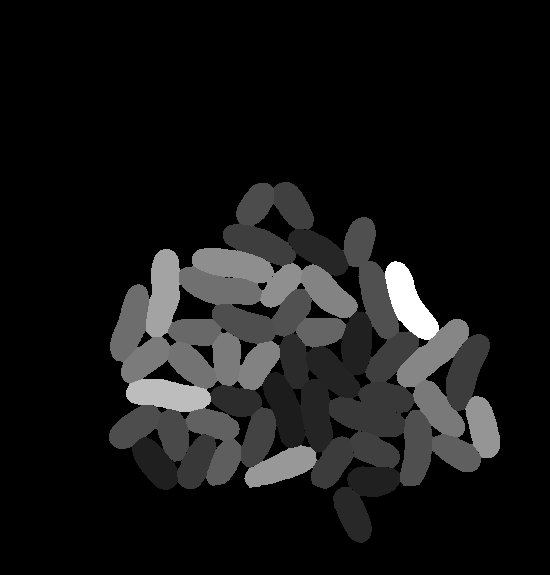}};
  \begin{scope}[x=(Bild.south east),y=(Bild.north west)]
    \node[color=white] at (0.075,0.925) {d)};
  \end{scope}
\end{tikzpicture}
\caption[]{Two consecutive phase contrast images $I_t,I_{t+1}$ of \textit{C.~glutamicum} (a-b), the corresponding standard deviation  $S_t$ (c), and the activity map (d).}
\label{fig:actmap}
\end{figure}
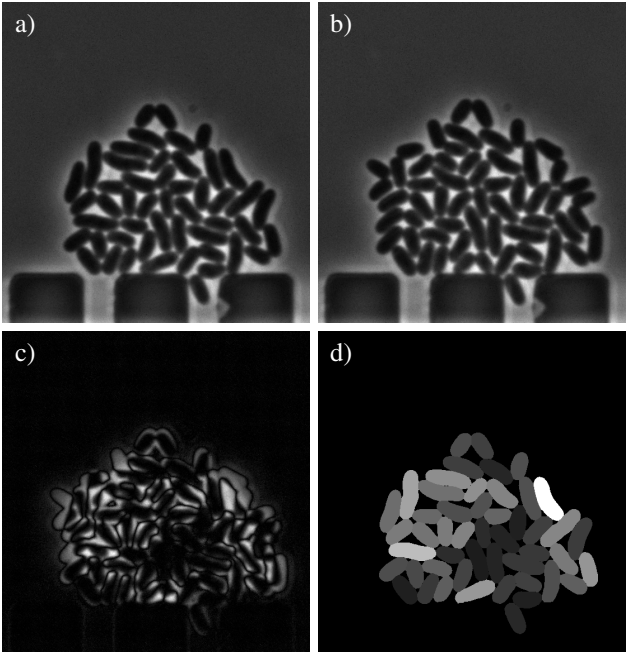

The proposed map may be beneficial to support manual tracking annotation, attracting the analyst's attention to the active (splitting, growing, or migrating) cells. Moreover, the AM allows automatic and straightforward \enquote{nearest neighbor} assignments for those cells that do not show any activity. 

One additional use of the activity map, that we don't explore here, is as an indicator to refine under-segmentation in the \enquote{non-active} regions by adding the \enquote{missed} non-active cell mask from the previous or the next frame.
Moreover, the AM can be used to refine the over-segmentation, e.g., segmenting pieces of the chip structure or air bubbles as cells, by concluding those objects that do not move during the entire cultivation time are unlikely cells.

Additionally, by being efficient and robust, the proposed AM may be useful  to adjust the cultivation conditions in real time, e.g., by indicating starving cell sub-colonies due to the lack of nutrition or other reasons.

\subsection{Gaussian activity-based heatmap as a linking measure}
Our target microorganism \textit{C.~glutamicum} is non-motile. Therefore, movement of a cell is expected only in the case of its own division or the division of the cells in its neighborhood. Moreover, the direction of the V-snap is hard or impossible to predict. Therefore, a linking measure to account for the mentioned facts is desired.

As a linking measure, the traditionally used Euclidean distance between cells' centers requires a globally set threshold to prevent non-physiological long-distance cell associations, and does not consider that some cells are less likely to divide than others.
To account for these characteristics, we propose a new linking measure, a new activity-based metric that restricts the daughter's search based on the mother's activity and, thus, offers fewer possible candidates compared to the Euclidean distance.

As an ingredient for the linking of the cells from $t$ to $t+1$, for each cell from $t$, we use a 2D Gaussian function $G$ centered at the center of mass of the $i$-th cell ($c_{i}$) and width $\sigma_i>0$:
\begin{equation}
    G_i(x,y) =\exp\left({-\tfrac{1}{2\sigma_i^2}((x-c_{x,i})^2 + (y-c_{y,i})^2)}\right).
\label{g}
\end{equation}
Here, $\sigma_i$ is proportional to the activity of the $i$-th cell ($a_i$), i.e, $\sigma_i=a_i/k$, with a scaling parameter $k$.
Since $G$ is strictly positive, but decaying exponentially, we reduce the number of possible candidates by treating $G$ below $0.01$ as zero. Unlike Euclidean distance thresholding, the proposed threshold is invariant to the spatial image resolution and does not need to be tuned. Fig.~\ref{result} illustrates the behavior of $G$ with two cells similar in size, but different in activity with $k=2.5$. This value of $k$ is used throughout this work.
 
 \begin{figure}[!htb]
\centering

\begin{tikzpicture}
  \node[anchor=south west,inner sep=0] (Bild) at (0,0)
    {\input{images/C.glutamicum_highlighted}};
  \begin{scope}[x=(Bild.south east),y=(Bild.north west)]
    \node[color=white] at (0.075,0.925) {a)};
  \end{scope}
\end{tikzpicture}
\begin{tikzpicture}
  \node[anchor=south west,inner sep=0] (Bild) at (0,0)
    {\input{images/C.glutamicum_highlighted2}};
  \begin{scope}[x=(Bild.south east),y=(Bild.north west)]
    \node[color=white] at (0.075,0.925) {b)};
  \end{scope}
\end{tikzpicture}
\\[0.75ex]
\begin{tikzpicture}
  \node[anchor=south west,inner sep=0] (Bild) at (0,0)
    {\begin{tikzpicture}

\definecolor{color0}{rgb}{0,0.75,0.75}

\begin{axis}[
hide axis,
width=0.38\textwidth,
axis equal image,
tick align=outside,
tick pos=left,
x grid style={white!69.0196078431373!black},
xmin=-0.5, xmax=549.5,
xtick style={color=black},
y dir=reverse,
y grid style={white!69.0196078431373!black},
ymin=-0.5, ymax=574.5,
ytick style={color=black}
]
\addplot graphics [includegraphics cmd=\pgfimage,xmin=-0.5, xmax=549.5, ymin=574.5, ymax=-0.5] {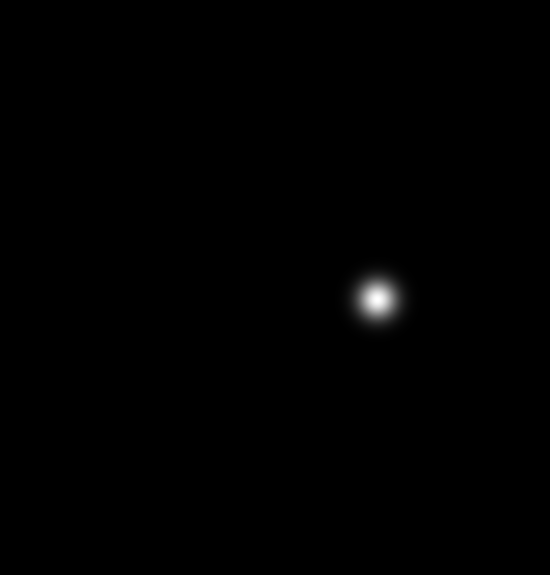};
\addplot [only marks, mark=square, mark size=0.05pt, color0]
table {%
378 334
378 333
379 335
380 336
381 337
382 337
383 337
384 338
385 338
386 338
387 339
388 338
389 338
390 337
391 337
392 336
393 335
394 335
395 334
396 333
397 332
398 331
398 330
398 329
398 328
398 327
398 326
398 325
398 324
397 322
398 323
397 321
397 320
397 319
397 318
396 317
396 316
395 315
395 314
394 313
394 312
393 311
393 310
393 309
392 308
392 307
391 306
391 305
391 304
390 303
390 302
390 301
390 300
389 299
389 298
389 297
388 296
388 295
388 294
388 293
387 292
387 291
387 290
387 289
386 288
386 287
386 286
386 285
386 284
385 283
385 282
385 281
385 280
385 279
384 278
384 277
384 276
384 275
384 274
384 273
383 272
382 271
382 270
381 269
380 268
380 267
379 266
378 265
377 264
376 263
375 262
374 262
373 262
372 261
371 261
370 261
369 261
368 261
367 261
366 262
365 262
364 263
363 264
362 265
361 265
360 266
359 267
359 268
359 269
359 270
359 271
359 272
359 273
359 274
359 275
359 277
358 276
359 278
359 279
359 280
359 281
359 282
359 283
360 284
360 285
360 286
360 287
361 288
361 289
361 290
361 291
361 292
361 293
361 294
362 295
362 296
362 297
362 298
362 299
363 300
363 301
363 302
363 303
364 304
364 305
364 306
365 307
365 308
365 309
366 310
366 311
365 312
366 313
367 314
368 315
368 316
369 317
369 318
370 319
370 320
371 321
371 322
371 323
372 324
372 325
372 326
373 326
374 327
374 328
375 329
376 330
376 331
377 332
378 334
};
\end{axis}

\end{tikzpicture}};
  \begin{scope}[x=(Bild.south east),y=(Bild.north west)]
    \node[color=white] at (0.075,0.925) {c)};
  \end{scope}
\end{tikzpicture}
\begin{tikzpicture}
  \node[anchor=south west,inner sep=0] (Bild) at (0,0)
    {\begin{tikzpicture}

\begin{axis}[
hide axis,
axis equal image,
width=0.38\textwidth,
tick align=outside,
tick pos=left,
x grid style={white!69.0196078431373!black},
xmin=-0.5, xmax=549.5,
xtick style={color=black},
y dir=reverse,
y grid style={white!69.0196078431373!black},
ymin=-0.5, ymax=574.5,
ytick style={color=black}
]
\addplot graphics [includegraphics cmd=\pgfimage,xmin=-0.5, xmax=549.5, ymin=574.5, ymax=-0.5] {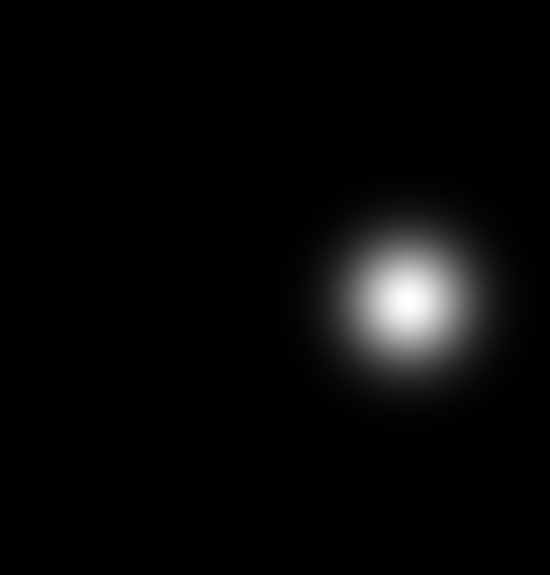};
\addplot [only marks, mark=square, mark size=0.05pt, red]
table {%
409 330
410 331
411 331
412 332
413 333
414 334
415 334
416 335
417 336
418 337
419 337
420 337
421 338
422 338
423 338
424 339
425 339
426 339
427 339
428 339
429 339
430 338
431 337
432 337
433 336
434 335
435 334
435 333
436 332
437 331
438 330
438 329
438 328
438 327
438 326
438 325
437 324
437 323
437 322
436 321
435 320
435 319
434 318
433 317
432 316
432 315
431 314
430 313
429 312
428 311
428 310
427 309
426 308
425 307
424 306
424 305
423 304
422 303
421 302
421 301
420 300
419 299
419 298
418 297
418 296
417 295
417 294
416 293
416 292
416 291
415 290
415 289
414 288
414 287
414 286
414 285
413 284
413 283
413 282
413 281
413 280
412 279
412 278
412 277
411 276
411 275
410 274
410 273
409 272
409 271
408 269
408 270
407 268
406 267
405 266
404 264
404 265
403 264
402 263
401 263
400 262
399 262
398 262
397 262
396 261
395 261
394 261
393 262
392 262
391 262
390 263
389 264
388 265
387 266
387 267
386 268
386 269
386 270
385 271
385 272
385 273
385 274
385 275
385 276
385 277
385 278
386 279
386 280
386 281
386 282
386 283
387 284
387 285
387 286
387 287
387 288
388 289
388 290
388 291
388 292
389 293
389 294
389 295
389 296
390 297
390 298
390 299
391 300
391 301
391 302
391 303
392 304
392 305
392 306
393 307
393 308
394 309
394 310
394 311
395 312
395 313
396 314
396 315
397 316
397 317
398 318
398 319
399 320
400 320
401 321
402 322
403 323
403 324
404 325
405 326
406 327
407 328
408 329
409 330
};
\end{axis}

\end{tikzpicture}};
  \begin{scope}[x=(Bild.south east),y=(Bild.north west)]
    \node[color=white] at (0.075,0.925) {d)};
  \end{scope}
\end{tikzpicture}

\caption[]{Frames at times $t$ (a) and $t+1$ (b) with highlighted cells: blue and red with relatively low and high activity in $t$ (see Fig. \ref{fig:actmap}) respectively. (b) shows the correct $t$ - $t+1$ assignment of the highlighted cells. (c) and (d) show the activity-based Gaussian maps $G$ or the respective cells-the linking measure of our proposed method. Note, that (c) and (d) were computed before the assignment and without using the tracking results shown in (b).}
\label{result}
\end{figure}

\subsection{Prioritization-based single assignment}
We propose to split the assignment in two stages: A prioritization-based single assignment of growing cells, followed by a combinatorial linear assignment of splitting cells and their daughters. The proposed assignment strategy is illustrated in Figure \ref{test}.
\begin{figure*}[ht!]

\includegraphics[width=.99\linewidth]{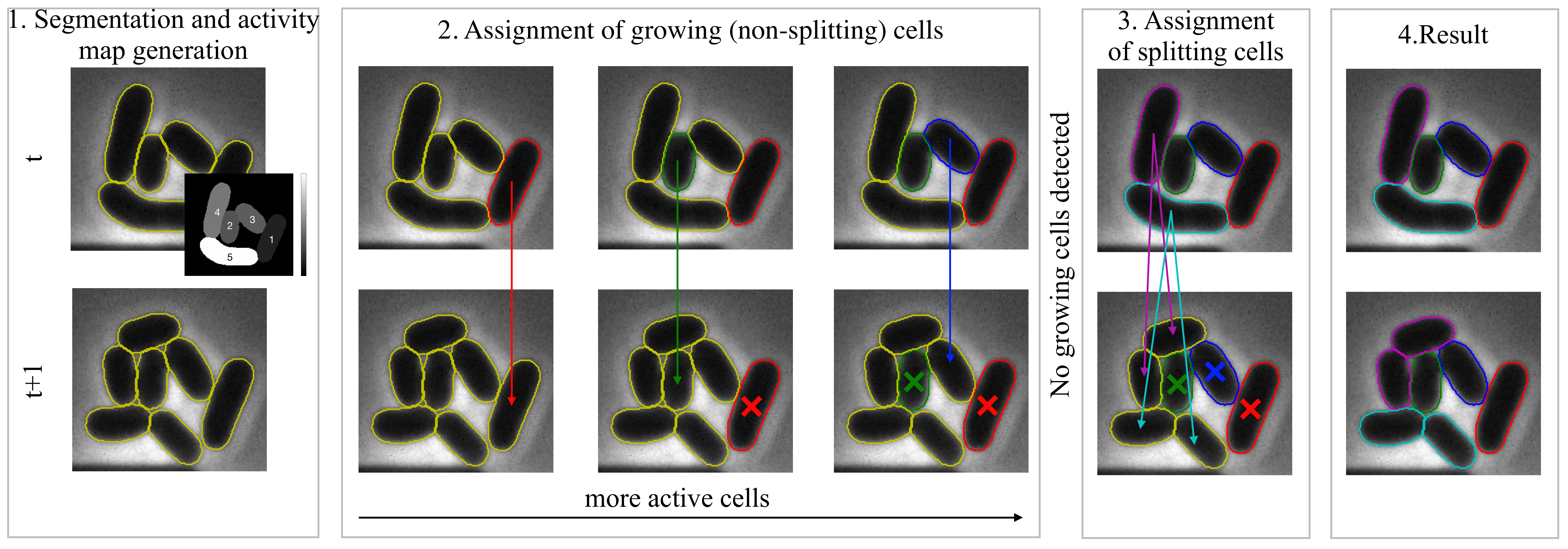}
\caption[]{The proposed tracking pipeline, which includes multi-object segmentation and activity map  generation (1) and the two-step assignment strategy (2 and 3), for cells from two consecutive frames. The assignment of growing cells starts with the cell with the lowest activity (2). Once all growing cells are assigned, the remaining cells from time $t$ and time $t+1$ participate in the combinatorial assignment (3).}
\label{test}
\end{figure*}
Before the assignment, we form two lists: the list of cells in frame $t$ and frame $t+1$, where the cells in $t$ are sorted in ascending order by activity ($a_i$).
We formulate the linking loss between the $i$-th cell of frame $t$ and the $j$-th cell of frame $t+1$ as follows:
\begin{equation}
    L(i,j)=-G_i(c_{x,j},c_{y,j}).
\end{equation}
Here, $G_i$ is the Gaussian map for the $i$-th cell of frame $t$, cf. Eq. \ref{g}, and $(c_{x,j},c_{y,j})\in\mathbb{R}^2$ is the center of mass of the $j$-th cell of frame $t+1$. 

We pair the cells (iterating over the mothers in the sorted order) with the minimal linking loss, and check if the criterion is satisfied. As a criterion for a \enquote{valid} link, we assume that cells are not shrinking in size, i.e., a cell from frame $t+1$ has to have at least as large area as its link from frame $t$. If the cell shrinking over time may be expected (e.g., under famine conditions), the criterion can be tailored, according to the specific behavior of the microorganism. If the criterion is satisfied, we remove the assigned cell from the list of possible candidates of frame $t$ and the list of frame $t+1$. 
After this initial assignment set, only the cells from frame $t$ that either split (in most cases) or would have been erroneously assigned to smaller cells and their possible daughters/growing \enquote{copies}, are left without assignment. These unassigned cells participate in a double linear assignment step.

\subsection{Linear assignment}
The linear sum assignment problem is known as minimum weight matching in bipartite graphs\cite{doi:10.1137/1.9781611972238}. A problem instance is described by a matrix $C$, where each matrix entry $C[i,j]$ is the cost (loss) of matching vertex $i$ of the first set (a \enquote{worker}, \enquote{mother} in our case) and vertex $j$ of the second set (a \enquote{job} or \enquote{daughter} in our case). The goal is to find a complete assignment of workers to jobs of minimal cost, i.e.,
\begin{equation}
\min_{X\in\mathcal{X}}\sum_{i}\sum_{j}C(i,j)X(i,j).
\end{equation}
Here, $\mathcal{X}$ is the set of boolean matrices $X$ with $\min(\text{num rows}(C),\text{num cols}(C))$ non-zero entries whose rows and columns sum to at most one and where, $X(i,j)=1$ if cell $i$ is assigned to cell $j$, and $X(i,j)=0$ otherwise. 

In case of $C$ being square, each row is assigned to exactly one column, and each column to exactly one row. Taking into account that one mother has exactly two daughters, we extend the matrix $C$, by doubling the number of rows (i.e.\ mothers), stacking two identical rectangular matrices on top of each other:

\begin{equation}
\label{eq:9}
C_{2m \times d}=
\begin{bmatrix}
\hat{C}_{m \times d}\\
\hat{C}_{m \times d}
\end{bmatrix}
\end{equation}
where
\begin{equation}
\label{eq:C_block}
\hat{C}_{m \times d}=
\begin{bmatrix}
L(1,1) & L(1,2) & \hdots & L(1,d)\\
L(2,1) & L(2,2) & \hdots & L(2,d)\\
\vdots&\vdots&\vdots&\vdots\\
L(m,1) & L(m,2) & \hdots & L(m,d)\\
\end{bmatrix}
\end{equation}
To solve the formulated problem, we use a modified Jonker-Volgenant algorithm without initialization \cite{Crouse2016}, implemented in \cite{2020SciPy-NMeth}.
This algorithm also solves a generalization of the classic assignment problem, where the cost matrix ($C$) is rectangular. If $C$ has more rows than columns, then not every row needs to be assigned to a column, and vice versa. In case of an odd number of daughters when $d>2m$, i.e., a new cell that was not present at frame $t$ appears in frame $t+1$, this cell is left unassigned. On the contrary, when $d<2m$, i.e., one of the daughters from frame $t+1$ disappears, the mother is only assigned one daughter.
The pseudocode of the proposed cell tracking strategy for two consecutive frames is shown as Algorithm \ref{algo}.
\RestyleAlgo{ruled}
\begin{algorithm}[ht]
	\caption{The proposed two-stage assignment strategy}
\KwData{Segmentation masks, Activity maps}
\KwResult{Cell pairs
}
\For{$t\leftarrow 0$ \KwTo $N-1$}{
\textit{Mothers} $\leftarrow$ [all \textit{cells} in $t$] \;
sort (\textit{Mothers}, ascending \textit{Activity})\;
\textit{Daughters} $\leftarrow$ [all \textit{cells} in $t+1$]\;
\textit{Pairs} $\leftarrow$ []\;
\For{every $\text{cell}_i$ in \textit{Mothers}
}{$\mathcal{L}$ $\leftarrow$ [] \;
\For{every $\text{cell}_j$ in \textit{Daughters}
}{$\mathcal{L}.append(L_{i,j})$}
\If{$\textit{area(cell}_i)>\textit{area(cell}_{\text{argmin}(\mathcal{L})})$}{
remove $cell_{\text{argmin}(\mathcal{L})}$ from \textit{Daughters}\;
remove $cell_i$ from \textit{Mothers}\;
$\textit{Pairs.append}([cell_i,cell_{\text{argmin}(\mathcal{L})}])$}}
$C$ $\leftarrow$ $L$(\textit{Mothers}, \textit{Daughters})\;
\textit{Pairs.append(linear\_sum\_assignment}($C$))\;
}
\label{algo}
\end{algorithm}
\section{Results}
\subsection{Ground truth generation}
To create a ground truth, we applied the proposed and a baseline approach (see below) to the segmented highest time-resolution dataset. In doing so, we carefully checked manually, and refined if needed, the obtained lineage. The original image sequences of the \textit{C.~glutamicum} and \textit{E.~coli} consist of 244 and 99 frames, respectively.
In order to evaluate our algorithm for different frame rates, i.e. 2, 4, 6, 8, 10, and 12 minutes for \textit{C.~glutamicum} and 5, 10, 15 minutes for \textit{E.~coli}, we down-sample our original datasets, respectively.

Additionally, we hope that the conducted frame rate sensitivity analysis of the algorithm for \textit{C.~glutamicum} and \textit{E.~coli} will provide insights to the biotechnologists about the choice of the frame rate for their cell cultivation experiments, offering a trade-off between tracking accuracy and microscopy settings.

\subsection{Comparison to a baseline approach}
We use the tracking accuracy (TRA) from the cell tracking challenge \cite{Ulman_2017} to evaluate the performance of the proposed algorithm. The TRA score is based on the acyclic-oriented graph matching measure \cite{Matula2015}, which penalizes the number of transformations needed to transform the predicted tracking graph into the ground truth tracking graph. The measure penalizes errors of the detection (false positives, false negatives) and errors concerning merged cells and tracking (missing links, wrong links, and links with wrong semantics).
Since our goal is to evaluate the tracking correctness only, we provide the ground truth segmentation for both baseline and the proposed approach and do not expect any detection-related errors.

As baseline algorithm, we use a graph-based cell tracking algorithm \cite{L_ffler_2021}, being ranked in the top three for the majority of the datasets of the cell tracking challenge.
The baseline algorithm offers segmentation refinement, such as untangling and false negative correction. Since, in our case, the tracking was performed on the ground truth segmentation with no need for refinement, we use two variants of the baseline approach: the algorithm with default parameter settings (\texttt{postprocessing\_key=None}), and without the segmentation refinement step (\texttt{postprocessing\_key='nd\_ns+l'}). In case of \emph{E.~coli} dataset evaluation, both settings lead to the same result, so we only report the value at default settings (Fig. \ref{plots}).

\begin{figure*}[ht!]
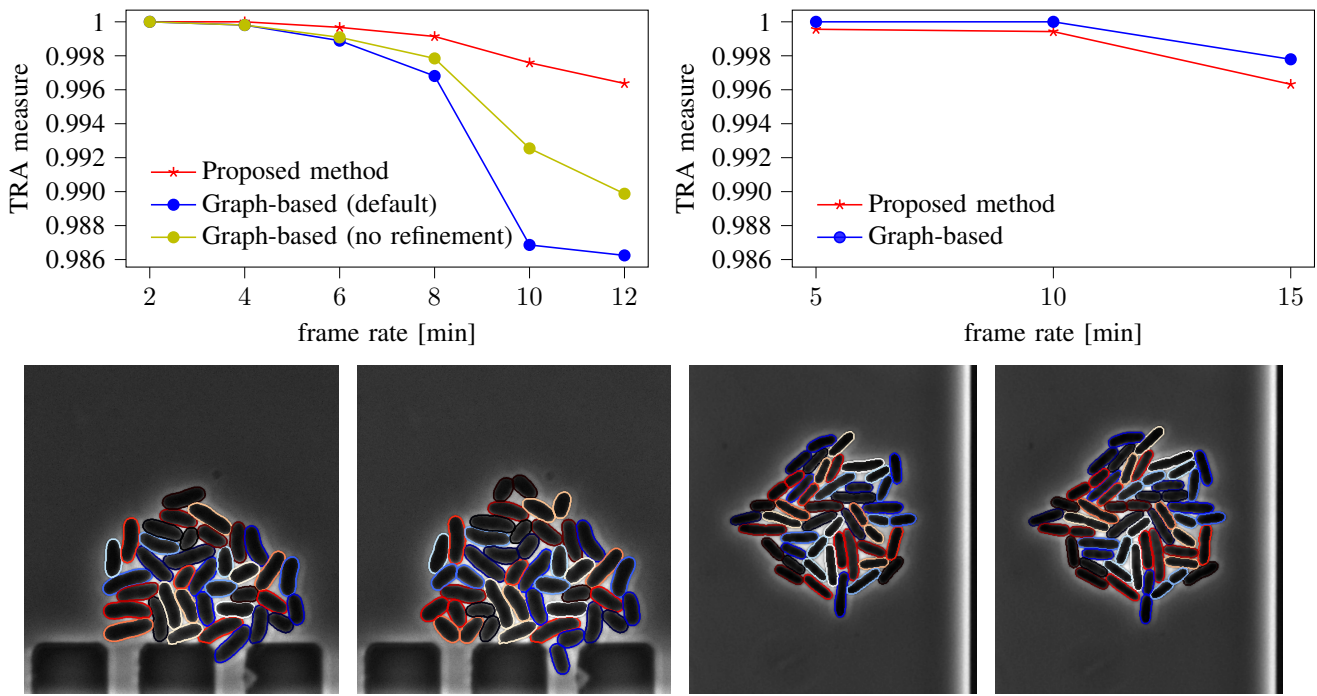

    \centering
\begin{tikzpicture}

\definecolor{color0}{rgb}{0.75,0.75,0}

\begin{axis}[
legend cell align={left},
width=0.47\textwidth,
height=5cm,
legend style={
fill=none,
  text opacity=1,
  at={(0.03,0.03)},
  anchor=south west,
  draw=none,
},
tick align=outside,
tick pos=left,
x grid style={white!69.0196078431373!black},
xlabel={frame rate [min]},
xmin=1.5, xmax=12.5,
xtick style={color=black},
y grid style={white!69.0196078431373!black},
ylabel={TRA measure},
ymin=0.98555725, ymax=1.00068775,
ytick={1,0.998,0.996,0.994,0.992,0.990,0.988,0.986},
yticklabels={1,0.998,0.996,0.994,0.992,0.990,0.988,0.986},
ytick style={color=black}
]
\addplot [semithick, red, mark=star, mark size=2, mark options={solid}]
table {%
2 1
4 1
6 0.999674
8 0.999138
10 0.997587
12 0.99637
};
\addlegendentry{Proposed method}
\addplot [semithick, blue, mark=*, mark size=2, mark options={solid}]
table {%
2 1
4 0.999809
6 0.99889
8 0.99681
10 0.986861
12 0.986245
};
\addlegendentry{Graph-based (default)}
\addplot [semithick, color0, mark=*, mark size=2, mark options={solid}]
table {%
2 1
4 0.999809
6 0.999086
8 0.997845
10 0.992546
12 0.989875
};
\addlegendentry{Graph-based (no refinement)}
\end{axis}

\end{tikzpicture}
\begin{tikzpicture}

\begin{axis}[
legend cell align={left},
width=0.47\textwidth,
height=5cm,
legend style={
  fill opacity=0.8,
  draw opacity=1,
  text opacity=1,
  at={(0.03,0.03)},
  anchor=south west,
  draw=none,
},
tick align=outside,
tick pos=left,
x grid style={white!69.0196078431373!black},
xlabel={frame rate [min]},
xmin=4.5, xmax=15.5,
xtick style={color=black},
xtick={5,10,15},
yticklabels={5,10,15},
y grid style={white!69.0196078431373!black},
ylabel={TRA measure},
ymin=0.98555725, ymax=1.00068775,
ytick style={color=black},
ytick={1,0.998,0.996,0.994,0.992,0.990,0.988,0.986},
yticklabels={1,0.998,0.996,0.994,0.992,0.990,0.988,0.986},
]
\addplot [semithick, red, mark=star, mark size=2, mark options={solid}]
table {%
5 0.999561
10 0.999424
15 0.996322
};
\addlegendentry{Proposed method}
\addplot [semithick, blue, mark=*, mark size=2, mark options={solid}]
table {%
5 1
10 1
15 0.997793
};
\addlegendentry{Graph-based}
\end{axis}

\end{tikzpicture}
\\[1ex]
\input{images/resglut12min35}
\input{images/resglut12min36}
\input{images/resglut15min31}
\input{images/resglut15min32}
\caption[]{Top row: A quantitative comparison of the proposed method vs. the baseline approach for different frame rates and two different microorganisms: C.~glutamicum (left) and E.~coli (right). The bottom row illustrates the tracking results of the proposed method for two consecutive frames of \textit{C.~glutamicum} (12 min) and \textit{E.~coli} (15 min), from left to right, respectively. Here, the color of the cell contour in $t$ is preserved for the same cell in $t+1$, as well as for both daughter cells, in the case of the cell's division.}
\label{plots}
\end{figure*}
The high TRA values in the obtained plots are explained by the fact that, the TRA measure's penalty factor (p) for the wrong detections: splitting operations (p=5), false negative vertices (p=10), false positive vertices (p=1)) is considerably higher, than for the wrong associations: redundant edges to be deleted (p=1), edges to be added (p=1.5) and edges with wrong semantics (p=1). Nevertheless, we use the TRA measure since it is known in the community, standardized and provides an unbiased comparison of the methods.

As shown in Fig.~\ref{plots}, the proposed method outperforms the baseline method when applied to a \textit{C.~glutamicum} dataset and is on par with the baseline approach applied to the \textit{E.~coli} dataset. 
The cross-correlation nature of the baseline approach and relatively simple (no snapping, the angle of daughters is preserved) division behavior of \textit{E.~coli}, explains the slightly better results of the method, whereas the proposed method handles the complex V-snapping division behavior of \textit{C.~glutamicum} better. Moreover, in case of the \textit{E.~coli} dataset, the septum formation as a part of the division process may not be captured in the phase contrast images. Therefore, the cell splitting events are sometimes not highlighted on the activity map. To overcome this, the fluorescence may be used to feature the septum formation~\cite{Sun_1998}.

\section{Conclusion}
We proposed a novel tracking-free mapping of active cells — the activity map. The activity map is useful as a standalone metric of cell activity, a tool to be used in tracking annotation software to attract the user's attention to the \enquote{changing} cells, or a building block for cell tracking. 
Additionally, we introduced a feature-free tracking approach that utilizes the activity map. The proposed tracking method consists of two steps: a prioritized single-cell assignment strategy of growing (non-splitting) cells and a combinatorial mother-daughter assignment of the dividing cells. We evaluated the proposed algorithm on datasets representing two important biotechnologically relevant rod-shaped microorganisms: \textit{C.~glutamicum}, with peculiar V-snapping division behavior, and \textit{E.~coli}, which divides uniformly by linear elongation. The numerical experiments show that the proposed tracking approach outperforms the baseline approach in \textit{C.~glutamicum} tracking and is on par with the baseline approach in \textit{E.~coli} tracking.
Additionally, we evaluate the algorithm's performance on datasets with different frame rates. We hope that the reported tracking quality metrics can provide insights to the biotechnologists and help them to choose the proper frame rate for their cultivation experiments.

Our source code is available on GitHub \url{https://github.com/kruzaeva/activity-cell-tracking}

\bibliographystyle{IEEEbib}
\bibliography{bib}

\end{document}